\documentclass[a4paper,10pt]{article}
\usepackage[utf8]{inputenc}
\usepackage{times}
\usepackage{geometry}
\usepackage{graphicx}
\usepackage{caption}
\usepackage{amsmath}
\usepackage{booktabs}
\usepackage{parskip}
\usepackage{setspace}
\usepackage{hyperref}
\usepackage{natbib}
\usepackage{amssymb}

\begin{document}

\title{\fontsize{24}{28}\selectfont ResLink: A Novel Deep Learning Architecture for Brain Tumor Classification with Area Attention and Residual Connections}
\author{\parbox{0.95\textwidth}{\centering Sumedha Arya, Nirmal Gaud \\ \href{mailto:aryasumedha06@gmail.com}{\url{aryasumedha06@gmail.com}}, \href{mailto:nirmal.gaud.ai@gmail.com}{\url{nirmal.gaud.ai@gmail.com}}}}
\maketitle

\begin{abstract}
{\fontsize{9}{11}\selectfont Brain tumors show significant health challenges due to their potential to cause critical neurological functions. Early and accurate diagnosis is crucial for effective treatment. In this research, we propose ResLink, a novel deep learning architecture for brain tumor classification using CT scan images. ResLink integrates novel area attention mechanisms with residual connections to enhance feature learning and spatial understanding for spatially rich image classification tasks. The model employs a multi-stage convolutional pipeline, incorporating dropout, regularization, and downsampling, followed by a final attention-based refinement for classification. Trained on a balanced dataset, ResLink achieves a high accuracy of 95\% and demonstrates strong generalizability. This research demonstrates the potential of ResLink in improving brain tumor classification, offering a robust and efficient technique for medical imaging applications.}
\end{abstract}

{\fontsize{9}{11}\selectfont \textbf{Keywords}---brain tumor, area attention, skip connections, deep learning, CT scan imaging}

\section{INTRODUCTION}
{\fontsize{10}{12}\selectfont Cancer is a life-threatening disease, caused by the abnormal and uncontrolled growth of harmful cells in the body \citep{anand2022cancer}. There are various forms of cancer. Among them, brain tumors pose a significant health challenge due to their potential to affect critical neurological functions. These are tumors that arise from abnormal brain cell growth. They can affect anyone, but men are slightly more at risk. The risk of certain tumors, such as glioblastoma, increases with age. The severity and prediction of brain tumors depend on several factors. Some of them are tumor type, location, size, and availability of effective treatment options.}

{\fontsize{10}{12}\selectfont Brain tumors are cancerous or non-cancerous, but can be life-threatening. Any abnormal growth can compress brain tissue, damage nerves, or block blood flow, leading to symptoms such as persistent headaches, seizures, cognitive problems, vision or hearing problems, and balance difficulties \citep{tandel2019review}. The exact cause of brain tumors is often unclear, but genetic mutations, environmental factors such as radiation exposure, and inherited conditions such as Neurofibromatosis may play a role. The diagnosis typically involves neurological exams, imaging tests such as MRI or CT scans, biopsies, and sometimes spinal taps to analyze cerebrospinal fluid. Early detection and treatment are crucial for effectively managing brain tumors.}

\begin{figure}[h]
\centering
\includegraphics[width=0.4\textwidth]{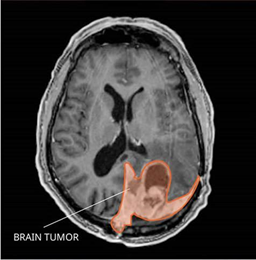}
\caption{Brain Tumor (\url{https://www.cancer.gov/rare-brain-spine-tumor/tumors/meningioma})}
\end{figure}

{\fontsize{10}{12}\selectfont Medical imaging plays a vital role in the early detection and diagnosis of brain tumors. In medical imaging, we use various tools and technologies to visualize the internal structures and functions of the human body for clinical analysis and medical diagnosis \citep{hussain2022modern}. The goal is to monitor, and treat diseases or conditions by creating detailed images that can be interpreted by medical professionals. These images help to detect abnormalities, and guided treatments. Magnetic Resonance Imaging (MRI) and Computed Tomography (CT) scans are widely utilized for this purpose.}

{\fontsize{10}{12}\selectfont In this research, we proposed a novel ResLink architecture based on deep learning. It integrates Area Attention with residual connections to enhance feature learning and spatial understanding. An attention mechanism is a technique that enables the model to selectively focus on the most relevant or informative regions in the CT scan images. The Area Attention Layer adaptively refined local feature regions, ensuring efficient information flow, while CNN blocks with residual connections preserve spatial integrity and gradient stability. The residual connections allow for smooth flow of gradients across layers in a deep learning architecture. ResLink employs a multi-stage convolutional pipeline, incorporating dropout regularization and downsampling, followed by a final attention-based refinement for classification. This architecture optimizes deep feature extraction, making it effective for spatially rich image classification tasks.}

{\fontsize{10}{12}\selectfont This paper is organized into five sections. Section 1 presents a comprehensive review of the literature on recent advances in brain tumor classification. Section 2 defines the problem statement, while Section 3 describes the proposed research methodology. Section 4 discusses the experimental results and their analysis, and Section 5 concludes the study with key findings and future directions.}

\section{LITERATURE REVIEW}
{\fontsize{10}{12}\selectfont The classification of brain tumors using advanced ML and DL techniques has gained priority in recent years. Several studies have shown significant progress in this domain, using various methodologies to achieve high accuracy in tumor classification. However, challenges such as limited dataset sizes, model complexity, and interpretability remain prevalent. In this section, we performed a detailed review of key studies in this field, highlighting their contributions and methodologies.}

{\fontsize{10}{12}\selectfont \citet{diaz2021deep} presented a novel deep learning approach for the classification and segmentation of brain tumors. They applied a multiscale convolutional neural network (CNN), which eliminated the need for preprocessing steps such as skull removal. Their model achieved a remarkable classification accuracy of 0.973 on a publicly available dataset of 3064 MRI slices from 233 patients.}

{\fontsize{10}{12}\selectfont \citet{labbaf2024brain} introduced an innovative approach to brain tumor classification using a Vision Transformer (ViT), with an improved Selective Cross-Attention (SCA) mechanism with Feature Calibration. They tried to solve the challenges of brain tumor classification by utilizing the ViT's strength to model long-range dependencies and combine multi-scale features. The model also calibrates features from different branches using feature calibration while SCA focuses on the most informative features. The suggested model, Cross ViT, achieves state-of-the-art performance, with an accuracy of 99.24\% and an F1-score of 99.23\% when used with a Stochastic Depth mechanism. The model has exceptional performance on metrics such as AUC (0.995), Recall, and Precision.}

{\fontsize{10}{12}\selectfont \citet{ranganathan2024brain} explored the application of machine learning to classify brain tumors using MRI images. They worked with two datasets: one to detect the presence or absence of tumors and another to classify tumor types. The researchers trained and tested different deep learning pre-trained models, including DenseNet, ResNet, EfficientNet, VGG-16, and Vision Transformer (ViT), with VGG-16 achieving the highest accuracy for tumor detection and DenseNet performing best in tumor classification. By combining these models, the authors further improved the accuracy of multiclass brain tumor classification. The study highlights how deep learning pre-trained models can boost medical imaging, leading to improved diagnostic accuracy, speed, and consistency, which can help healthcare professionals in planning treatment.}

{\fontsize{10}{12}\selectfont \citet{liu2024deep} explored the application of deep learning for the classification of brain tumor using MRI images only. They highlighted the importance of accurate and timely diagnosis of brain tumors in improving patient outcomes. The research utilized four pre-trained models: MobileNetV2, ResNet-18, EfficientNet-B0, and VGG16. Although the accuracies of these models ranged from 0.8445 to 0.9497, the new model called MobileNet-BT, based on MobileNetV2, achieved a higher accuracy of 0.9924. Their study showed the importance of customizing deep learning architectures for tasks like medical image classification.}

{\fontsize{10}{12}\selectfont \citet{liu2024brain} looked at the issue of predicting 1p/19q co-deletion status in low-grade gliomas using an MRI-based CNN architecture, an important factor for deciding on treatment and improving patient outcomes. The authors pointed out the disadvantages of using transfer learning with pre-trained models like InceptionV3, VGG16, and MobileNetV2. While these models are effective in general image classification, they often include irrelevant weights for medical imaging, leading to inaccurate diagnostic results. To overcome this issue, the researchers suggested a custom CNN model built from scratch, incorporating convolution stacking, dropout, and fully connected layers to reduce overfitting and improve performance. The model was trained using a three-fold cross-validation method, and Gaussian noise was added to address dataset imbalance. The proposed model showed impressive results, achieving a 96.37\% F1-score, 97.46\% precision, and 96.34\% recall on a validation set of 1p/19q co-deletion and non-co-deletion images, surpassing fine-tuned pre-trained models. This study emphasizes the need to create specialized, lightweight, and dependable models for medical imaging tasks, particularly when working with limited and imbalanced datasets.}

{\fontsize{10}{12}\selectfont \citet{pandiyaraju2024integrated} introduced an integrated deep learning framework for brain tumor localization, segmentation, and classification using MRI images. Their work focused on the importance of early and accurate diagnosis to address severe neurological problems. The authors suggested a multi-stage approach, beginning with an improved LinkNet framework. This framework included a VGG19-inspired encoder and spatial and graph attention mechanisms for accurate tumor localization and feature extraction. In the segmentation task, the SeResNet101 model integrated into the LinkNet architecture reached an impressive IoU score of 96\%. For the classification task, it combined SeResNet152 with an Adaptive Boosting classifier, achieving an accuracy of 98.53\%. The proposed framework comprised advanced methods such as Squeeze-and-Excitation (SE) blocks and attention mechanisms to improve feature sensitivity and focus on critical tumor areas. The research showed the transformative impact of integrating CNNs, SE blocks, and attention mechanisms in medical imaging, paving the way to improve patient care.}

{\fontsize{10}{12}\selectfont \citet{oh2024machine} explored the application of ML and DL models for brain tumor detection and classification using MRI images. They highlighted the importance of early brain tumor detection and accurate diagnosis. They used various models, including linear, logistic, and Bayesian regressions, decision trees, random forests, perceptrons, convolutional neural networks (CNNs), recurrent neural networks (RNNs), and long short-term memory (LSTM) networks. Among these models, CNNs emerged as the best-performing model, excelling in both binary and multi-class classification tasks. The study highlighted the importance of CNNs in extracting spatial features and learning from complex MRI data, achieving high accuracy, precision, recall, and F1 scores.}

{\fontsize{10}{12}\selectfont \citet{paul2024efficient} focused on the critical problem of identifying brain tumors using MRI images, emphasizing the importance of rapid and accurate diagnosis. The authors proposed a Convolutional Neural Network (CNN)-based architecture for detecting tumors, which achieved an impressive accuracy of 99.17\%. The CNN model extracted features, while machine learning (ML) models such as KNN, Logistic Regression, SVM, Random Forest, Naive Bayes, and Perceptron were used for classification. The research highlighted the superiority of deep learning methods, particularly CNNs, in medical imaging tasks, where even small diagnostic mistakes can have severe consequences.}

{\fontsize{10}{12}\selectfont \citet{ketabi2024tumor} presented a new contrastive learning (CL) framework that combines 3D MRI scans, radiology reports, and tumor location information. This approach aimed to enhance the clarity and efficiency of pediatric brain tumor diagnosis, particularly for low-grade gliomas (pLGG), the most common type of brain tumor in children. Although CNNs are successful in brain tumor diagnosis, their clinical adoption has been limited due to a lack of explainability, as the features driving predictions are often ambiguous to radiologists. To address this, the authors proposed a multimodal CL architecture that leverages associations between MRI scans, radiology reports, and tumor location data to learn more useful and clinically relevant features. When applied to the downstream task of pLGG genetic marker classification, it achieved an AUC of 87.7\% and demonstrated better explainability, with Dice scores of 31.1\% (2D) and 15.8\% (3D) between the model’s attention maps and manual tumor segmentations.}

{\fontsize{10}{12}\selectfont \citet{mohammadi2024enhancing} explored advanced machine learning techniques, including TrAdaBoost and multi-classifier deep learning algorithms, to improve brain tumor classification through MRI images. The authors addressed the issue of accurately identifying and characterizing brain tumors by integrating state-of-the-art models such as Vision Transformer (ViT), Capsule Neural Network (CapsNet), and convolutional neural networks (CNNs) like ResNet-152 and VGG16 within a multi-classifier framework. They introduced a new decision template to combine the outputs of these models, improving classification accuracy and robustness. The research used the BraTS2020 dataset, augmented with the "Brain Tumor MRI Dataset," to improve generalization and training efficiency. Their results demonstrated high accuracy in distinguishing tumor from non-tumor images. The integration of TrAdaBoost with deep learning models highlighted the feasibility of automated diagnostic tools in improving patient outcomes.}

{\fontsize{10}{12}\selectfont \citet{karagoz2024residual} presented a self-supervised learning (SSL) model based on a Residual Vision Transformer (ResViT) to classify brain tumors, responding to the challenge of limited annotated MRI datasets. It employed a two-stage approach: first, pre-training a ResViT model for MRI synthesis, utilizing both local features through convolutional neural networks (CNNs) and global features via Vision Transformers (ViTs). In the second stage, the ResViT-based classifier was fine-tuned for tumor classification, with synthetic MRI images used to balance the training dataset and improve outcomes. The model was tested on public datasets (BraTS 2023, Figshare, and Kaggle) and outperformed state-of-the-art models in both MRI synthesis and classification tasks, achieving accuracies of 90.56\% (BraTS T1 sequence), 98.53\% (Figshare), and 98.47\% (Kaggle). The study demonstrated the effectiveness of integrating SSL, hybrid CNN-ViT architectures, fine-tuning, and data augmentation to overcome dataset limitations and improve generalization.}

{\fontsize{10}{12}\selectfont \citet{khan2024comparative} conducted a comparative analysis of resource-efficient convolutional neural network architectures for brain tumor classification, focusing on the trade-off between accuracy and computational efficiency. The authors evaluated their custom CNN model against pre-trained models like ResNet-18 and VGG-16 using two publicly available datasets: Br35H: Brain Tumor Detection 2020 and Brain Tumor MRI Dataset. The customized CNN architecture, with lower complexity, achieved competitive performance, with accuracies of 98.67\% (Br35H) and 99.62\% (Brain Tumor MRI Dataset) in binary classification tasks, and 98.09\% in multi-class classification. ResNet-18 and VGG-16 maintained high accuracy, but custom CNN offered significant benefits in computational efficiency, including faster training and inference times. The study also explored few-shot learning scenarios, showcasing the robustness and adaptability custom CNN to limited data.}

{\fontsize{10}{12}\selectfont \citet{rivera2024ensemble} suggested an ensemble deep learning framework for brain tumor classification and synthesis. They utilized state-of-the-art architectures such as nn-UNet, Swin-UNet, and U-Mamba to enhance accuracy and image analysis in neuroimaging. They highlighted that nn-UNet showed superior performance in individual evaluations, while U-Mamba stood out with shorter runtime. The ensemble method, along with effective post-processing strategies, significantly improved the robustness and accuracy of tumor classification, particularly in reducing false positives across diverse tumor types. The study also explored advanced techniques like Gated Linear Attention and Fourier Blocks in the MA3T-F model, which reduced computational complexity and improved global contextual information capture, allowing for faster generalization. Additionally, the integration of WGANs and GrokFast algorithms further improved model performance.}

{\fontsize{10}{12}\selectfont \citet{nafi2024diffusion} explored how diffusion models can address the problem of data scarcity in medical imaging by generating synthetic datasets for training convolutional neural networks (CNNs). Their study focused on three domains: Brain Tumor MRI, Acute Lymphoblastic Leukemia (ALL), and SARS-CoV-2 CT scans. In their analysis, they found that CNNs trained on synthetic data performed well on real-world test datasets, with a test accuracy of 91.38\%. They also conducted Local Interpretable Model-Agnostic Explanations (LIME) analysis to support that these models focused on relevant image features. However, the authors emphasized the need for further research to fine-tune hyperparameters, evaluate dataset diversity, and assess performance with new datasets.}

{\fontsize{10}{12}\selectfont \citet{alam2024enhancing} investigated enhancing transfer learning for multilabel medical image classification, specifically focusing on brain tumor classification and diabetic retinopathy stage detection. The authors assessed five pre-trained models: MobileNet, Xception, InceptionV3, ResNet50, and DenseNet201, on two datasets: Brain Tumor MRI and APTOS 2019. Although transfer learning models showed strong performance in brain tumor classification, their ability to detect diabetic retinopathy was limited by class imbalance. To address this, the authors combined the Synthetic Minority Over-sampling Technique (SMOTE) with transfer learning and machine learning techniques, resulting in significant improvements in accuracy, recall, and specificity. The research highlighted the importance of combining transfer learning with resampling techniques to manage imbalanced datasets and enhance classification performance.}

{\fontsize{10}{12}\selectfont \citet{ganguly2025efficient} introduced a novel lightweight CNN architecture for classifying brain tumors using MRI images accurately and efficiently. The proposed model used separable convolutions and squeeze-and-excitation (SE) blocks to improve feature extraction while maintaining computational efficiency. Additionally, they incorporated batch normalization, global average pooling, and dropout layers to prevent model overfitting. Experimental results showed the model performed well, achieving a validation accuracy of 99.22\% and a test accuracy of 98.44\%, outperforming existing models by 0.5\% to 1.0\% in accuracy and 1.5\% to 2.5\% in loss reduction. This research highlighted the model’s robustness and generalization capabilities across diverse brain tumor types, setting a new benchmark in the field.}

{\fontsize{10}{12}\selectfont \citet{hasan2025cnn} offered a thorough method to automate brain tumor detection and classification using advanced deep learning techniques with MRI data. They tackled issues in medical imaging, such as noise and incomplete images, by employing an anisotropic diffusion filter for denoising. Further, the Synthetic Minority Over-sampling Technique (SMOTE) was used for data augmentation and balancing. They evaluated several Convolutional Neural Network (CNN) architectures, including ResNet152V2, VGG, Vision Transformer (ViT), and EfficientNet, on a publicly available Brain Tumour Classification (MRI) dataset consisting of 3,264 scans. EfficientNet emerged as the top-performing model, with an accuracy of 98\%, outperforming other architectures in terms of precision, recall, and F1-score.}

{\fontsize{10}{12}\selectfont \citet{abdusalomov2023brain} developed an advanced deep learning system for brain tumor detection in MRI images. They used a state-of-the-art model, YOLOv7, along with techniques like transfer learning, fine-tuning, and tools like CBAM and SPPF+ to improve tumor detection and classification. Their model achieved an impressive 99.5\% accuracy, excelling in identifying gliomas, meningiomas, and pituitary tumors despite variations in size, texture, or location. However, the authors noted that further research is needed to test the system on smaller tumors and a wider variety of cases.}

{\fontsize{10}{12}\selectfont \citet{irmak2021multi} developed custom deep learning architectures for brain tumor detection, multi-classification, and grading gliomas into three severity levels (Grade II, III, IV). Their CNN models achieved accuracies of 99.33\%, 92.66\%, and 98.14\%, outperforming pretrained models like AlexNet and ResNet-50. Using grid search for fine-tuning and large public datasets, the system proved effective for early and accurate diagnosis.}

{\fontsize{10}{12}\selectfont \citet{nassar2023robust} developed an ensemble deep learning model for brain tumor classification using MRI images, combining five CNN-based models with a majority voting method, achieving 99.31\% accuracy. The system is fast and reliable but requires more diverse data for less common tumors like meningioma. The authors suggested its applicability to other medical tasks, such as detecting lung cancer, skin lesions, or breast cancer.}

{\fontsize{10}{12}\selectfont \citet{albalawi2024integrated} built a deep learning model for brain tumor classification using MRI images, combining federated learning and transfer learning with a modified VGG16 model. Trained on a diverse dataset, it achieved 98\% accuracy in identifying glioma, meningioma, no tumor, and pituitary tumor types, offering a faster, more accurate, and privacy-preserving diagnostic tool.}

{\fontsize{10}{12}\selectfont \citet{zaineldin2023brain} developed an advanced deep learning model, BCM-CNN, for brain tumor detection and classification, using CNNs optimized with ADSCFGWO for fine-tuning. It achieved 99.98\% accuracy on the BRaTS 2021 dataset, though it has high computational complexity and processing time. Future improvements aim to reduce complexity and expand to larger datasets for tumor prediction.}

\section{PROBLEM STATEMENT}
{\fontsize{10}{12}\selectfont Despite the significant advancements in ML and DL algorithms for brain tumor classification, existing approaches still face challenges in computational complexity, and robustness across diverse datasets. In this section, we identified various problem statements based on the literature review as follows:}

\begin{enumerate}
\item Challenges in Brain Tumor Classification to achieve high accuracy with less computational resources.
\item Limitations of Pre-trained Models in Medical Imaging due to use of irrelevant fine tuning of weights.
\item Handling Imbalanced and Limited Datasets in Medical Imaging.
\end{enumerate}

{\fontsize{10}{12}\selectfont In this research, we focused on reducing model computational complexity, without the use of a pre-trained model, and achieving high accuracy with a balanced dataset.}

\section{RESEARCH METHODOLOGY}
{\fontsize{10}{12}\selectfont In this section, we explain the entire research methodology used in the study. It involves several steps, including data collection, preprocessing, model development, and training.}

\subsection{Dataset}
{\fontsize{10}{12}\selectfont The dataset is collected from Kaggle, containing brain tumor CT scan images categorized into two classes: "Healthy" and "Tumor." It is a benchmark dataset for brain tumor image classification tasks.}

\begin{figure}[h]
\centering
\includegraphics[width=0.4\textwidth]{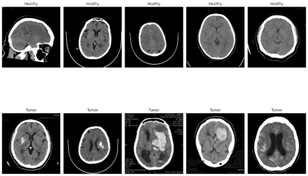}
\caption{Brain Tumor Dataset Images}
\end{figure}

\subsection{Exploratory Data Analysis}
{\fontsize{10}{12}\selectfont Basic exploratory data analysis (EDA) is performed to understand the dataset's structure, including checking duplicates and missing values.}

\begin{figure}[h]
\centering
\includegraphics[width=0.4\textwidth]{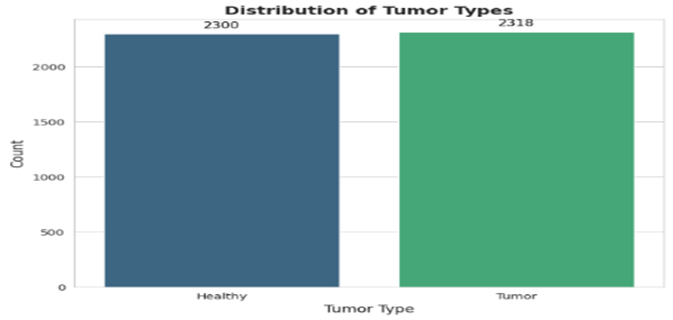}
\caption{Distribution of Tumor Types}
\end{figure}

\subsection{Data Preprocessing}
{\fontsize{10}{12}\selectfont The categorical labels ("Healthy" and "Tumor") are encoded into numerical values (0 and 1) using LabelEncoder for compatibility with machine learning models. The dataset is resampled using RandomOverSampler to address class imbalance, ensuring an equal number of samples for both classes.}

\subsection{Dataset Splitting}
{\fontsize{10}{12}\selectfont The dataset is split into training (80\%), validation (10\%), and test (10\%) subsets. To ensure that each split preserves the original class distribution, stratified sampling is employed during the splitting process.}

\subsection{Image Preprocessing}
{\fontsize{10}{12}\selectfont Images are resized to a uniform size of (224, 224) pixels and normalized by rescaling pixel values to the range [0, 1]. ImageDataGenerator is used to create data generators for training, validation, and testing. The generators handle image loading, preprocessing, and batching during model training and evaluation.}

\subsection{ResLink Architecture}
{\fontsize{10}{12}\selectfont A custom Convolutional Neural Network (CNN) model architecture, named ResLink, is built. It includes an Area Attention Mechanism, a custom layer (AreaAttentionLayer) to focus on specific regions in an image; multiple convolutional blocks with Batch Normalization and Dropout for regularization; and Dense layers for final classification. The model uses binary cross-entropy loss and the Adam optimizer as a gradient descent optimization algorithm.}

\subsubsection{Input and Stem Layer:}
{\fontsize{10}{12}\selectfont Given an input image \( X \in \mathbb{R}^{H \times W \times C} \), the stem performs convolution and downsampling:
\begin{equation}
F0 = \text{MaxPool}(\text{ReLU}(\text{BN}(W_s * X))),
\end{equation}
where \( W_s \) is a 7x7 convolution kernel with stride 2, * denotes convolution, BN is batch normalization, and ReLU is the activation function.}

\subsubsection{Area Attention Layer:}
{\fontsize{10}{12}\selectfont Let the feature map be \( F \in \mathbb{R}^{h \times w \times c} \), divided into non-overlapping spatial areas of size \( a \times b \). Each area region \( A_{i,j} \in \mathbb{R}^{a \times b \times c} \) is processed to generate an attention weight:
\begin{equation}
\alpha_{i,j} = \text{sigmoid}(W_2 * \text{BN}(W_1 * A_{i,j})),
\end{equation}
where \( W_1 \) is a 1x1 convolution, \( W_2 \) is a 3x3 convolution, and sigmoid is the activation function. The attention map is reshaped back and multiplied element-wise with the original features:
\begin{equation}
F_{\text{att}} = F \odot \alpha,
\end{equation}
where \( \odot \) denotes elementwise (Hadamard) product.}

\subsubsection{Residual CNN Block:}
{\fontsize{10}{12}\selectfont Given an input feature map \( F \in \mathbb{R}^{h \times w \times c} \), the block generates:
\begin{equation}
G1 = \text{ReLU}(\text{BN}(W_{b1} * F)),
\end{equation}
\begin{equation}
G2 = \text{BN}(W_{b2} * G1),
\end{equation}
where \( W_{b1} \) and \( W_{b2} \) are 3x3 convolution kernels. The shortcut connection is:
\begin{equation}
S = \begin{cases} 
\text{BN}(W_s * F), & \text{if } c \neq \text{number\_of\_filters} \\
F, & \text{otherwise}
\end{cases}
\end{equation}
Then the residual connection output is:
\begin{equation}
H = \text{ReLU}(G2 + S)
\end{equation}
(When attention is enabled, \( F \) is replaced by \( F_{\text{att}} \) before these convolutions.)}

\subsubsection{Downsampling Layer:}
{\fontsize{10}{12}\selectfont To reduce spatial dimension:
\begin{equation}
F_{\text{down}} = \text{ReLU}(\text{BN}(W_d * H)),
\end{equation}
where \( W_d \) is a 3x3 convolution with stride 2.}

\subsubsection{Final Attention and Classification:}
{\fontsize{10}{12}\selectfont Final Area Attention is applied:
\begin{equation}
F_{\text{final}} = \text{AreaAttention}(F_{\text{down}})
\end{equation}
Global Average Pooling computes the k-th channel of the output vector as:
\begin{equation}
z_k = \frac{1}{h \times w} \sum_{i=1}^{h} \sum_{j=1}^{w} F_{\text{final}}^{i,j,k}
\end{equation}
\begin{equation}
z' = \text{Dropout}(z)
\end{equation}
The final classifier predicts:
\begin{equation}
\hat{y} = \begin{cases} 
\text{sigmoid}(W \times z' + b), & \text{for binary classification} \\
\text{softmax}(W \times z' + b), & \text{for multi-class classification}
\end{cases}
\end{equation}}

\begin{figure}[h]
\centering
\includegraphics[width=0.4\textwidth]{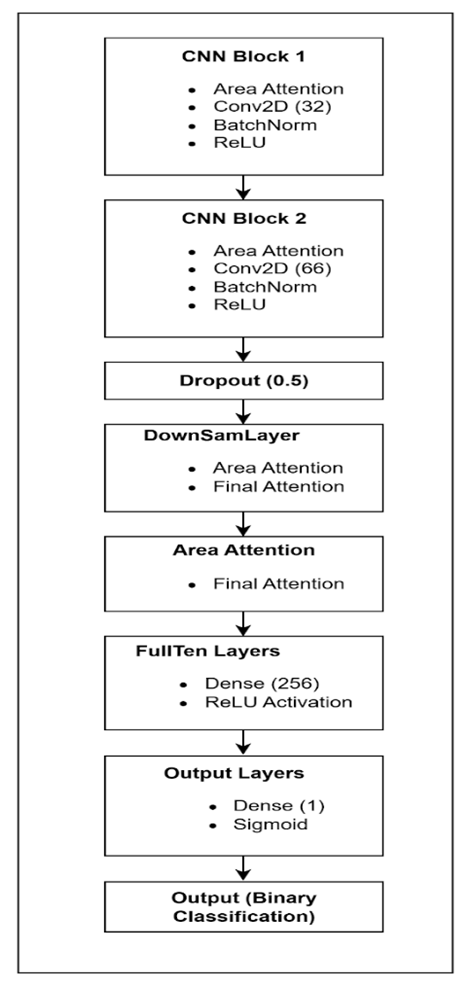}
\caption{ResLink Architecture}
\end{figure}

\subsection{Model Training}
{\fontsize{10}{12}\selectfont The model is trained for 5 epochs using the training data generator. It is performed on a GPU for faster computation. Training and validation accuracy/loss are monitored to assess model performance during training.}

\section{RESULT ANALYSIS}
{\fontsize{10}{12}\selectfont The proposed ResLink model demonstrated rapid improvement in accuracy within the initial epochs. In Epoch 1, the training accuracy was 77.68\%, with a relatively high loss of 0.7382, but the validation accuracy was already at 92.67\%, indicating a promising start. By Epoch 2, the model significantly improved, reaching 95.25\% training accuracy, with validation accuracy at 94.83\%.}

\begin{figure}[h]
\centering
\includegraphics[width=0.4\textwidth]{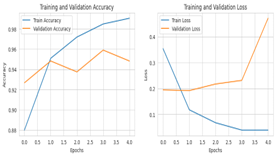}
\caption{Training and Validation Accuracy and Loss Plots}
\end{figure}

{\fontsize{10}{12}\selectfont From the confusion matrix and classification report, ResLink shows strong generalization capabilities. The model achieved high precision (0.97) and recall (0.92) for class 1- Healthy, confirming effective feature learning. The macro-average F1-score of 0.95 highlights balanced performance across both classes, with minimal misclassifications. This suggests that the Area Attention and ResLink mechanisms are effectively capturing important patterns in the data.}

\begin{figure}[h]
\centering
\includegraphics[width=0.4\textwidth]{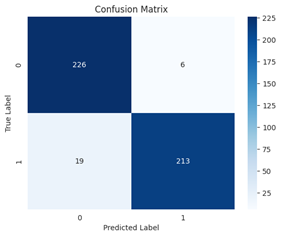}
\caption{Confusion Matrix Healthy vs Tumor}
\end{figure}

{\fontsize{10}{12}\selectfont To further improve model performance, increasing dataset diversity through augmentation or additional data collection could enhance generalization. Additionally, experimenting with learning rate scheduling and fine-tuning dropout rates could refine ResLink's ability to classify unseen data more accurately. Finally, our proposed model achieves high classification accuracy (95\%), demonstrating its effectiveness in using Area Attention and ResLink mechanisms.}

\section{CONCLUSION}
{\fontsize{10}{12}\selectfont In conclusion, brain tumors are a serious health issue that can affect anyone, with risks increasing with age. Early detection and treatment for brain tumors are crucial. In this research, ResLink, a novel model, showed promising results, achieving high accuracy and effectively distinguishing between healthy and tumorous images. While there is scope for improvement, such as increasing dataset diversity and fine-tuning the model, ResLink demonstrates the potential of deep learning in medical imaging. This research highlights the importance of innovative technologies in improving brain tumor diagnosis and treatment, ultimately helping patients and healthcare professionals.}

{\fontsize{8}{10}\selectfont
\bibliographystyle{plainnat}
\bibliography{references}
}

\end{document}